\documentclass[10pt]{article} 
\usepackage[accepted]{tmlr}

\usepackage{amsmath,amsfonts,bm}

\def\eqref#1{equation~\ref{#1}}

\def\1{\bm{1}}

\DeclareMathAlphabet{\mathsfit}{\encodingdefault}{\sfdefault}{m}{sl}
\SetMathAlphabet{\mathsfit}{bold}{\encodingdefault}{\sfdefault}{bx}{n}

\usepackage{hyperref}
\usepackage{url}
\usepackage{graphicx}
\usepackage{dirtytalk}
\usepackage{enumitem}

\newcommand\blfootnote[1]{%
  \begingroup
  \renewcommand\thefootnote{}\footnote{#1}%
  \addtocounter{footnote}{-1}%
  \endgroup
}

\definecolor{ferngreen}{rgb}{0.31, 0.47, 0.26}
\definecolor{carrotorange}{rgb}{0.93, 0.57, 0.13}

\title{Understanding Causality with Large Language Models:\\ Feasibility and Opportunities}

\author{\name Cheng Zhang, Stefan Bauer, Paul Bennett, Jiangfeng Gao, Wenbo Gong, Agrin Hilmkil, Joel Jennings, Chao Ma, Tom Minka, Nick Pawlowski, James Vaughan}

\begin{document}

\maketitle

\begin{abstract}
We assess the ability of large language models (LLMs) to answer causal questions by analyzing their strengths and weaknesses against three types of causal question. We believe that current LLMs can answer causal questions with existing causal knowledge as combined domain experts. However, they are not yet able to provide satisfactory answers for discovering new knowledge or for high-stakes decision-making tasks with high precision. We discuss possible future directions and opportunities, such as enabling explicit and implicit causal modules as well as deep causal-aware LLMs. These will not only enable LLMs to answer many different types of causal questions for greater impact but also enable LLMs to be more trustworthy and efficient in general.
\blfootnote{Correspond to: cheng.zhang@microsoft.com; the rest of the authors are ordered by family name initials.\\

Claim: This commentary are based only on publicly available information such as public APIs and technical articles by March 2023. }
\end{abstract}

\section{Introduction}
The development of large language models \cite{brown2020language,chatgpt, openai2023gpt4, ouyang2022training, radford2018improving, radford2019language, thoppilan2022lamda, vaswani2017attention} has been extremely rapid. Recently, ChatGPT \cite{chatgpt,  openai2023gpt4,ouyang2022training} has disrupted many domains, such as search, AI-aided content generation and editing, and is making great strides towards artificial general intelligence (AGI) \cite{bubeck2023sparks}. On the other hand, LLMs have been shown to have some limitations in formal reasoning \cite{bubeck2023sparks, mahowald2023dissociating,wolfram}.

One of the most commonly asked question types are causal questions (several examples of such questions are given in Section \ref{sec:limtation}). As humans actively interact with the environment, causal questions are fundamental to our nature. Besides its ultimate importance, understanding causal questions will likely be important for AGI and beyond. Answering such questions requires a combination of symbolic reasoning and high precision prediction; efficient interaction with humans to understand and verify assumptions; discovery of unknown knowledge; and interfacing of real-world physical interactions with safety awareness. 
Thus, in this article, we discuss LLMs in terms of their causal reasoning capabilities and the opportunities going forward. 

In particular, we want to discuss and examine to what extent LLMs can be used to answer questions of causal nature.
We will initially focus on current LLMs, a type of auto-regressive generative model \cite{brown2020language, radford2019language}, characterized by the next token generation-based training. ChatGPT \cite{chatgpt, openai2023gpt4,ouyang2022training} introduces a new training paradigm where human feedback is utilized to enhance the alignment to the human objective regarding usefulness. The current usage of such human interventions is in its infancy, and we will discuss its potential in Section \ref{sec:opportunity}.
Using causal reasoning as an example, our discussion perhaps generalizes beyond causal questions towards reasoning skills.    

 LLMs have obtained impressive language skills, providing a generic and natural way for human to interact with AI \cite{mahowald2023dissociating}. It has also obtained a large volume of knowledge about the world
which in many cases appear as a "conservative team of experts" \cite{guo2023close}.

Most impressively, it demonstrated strong understanding beyond memorization in a large range of scenarios including correct tool use, basic mathematical reasoning and understanding the mental state of human and agents \cite{bubeck2023sparks}.
LLMs can answer causal questions that are rooted in common knowledge or using known tools (as \ref{item:type1}). 
However, when deep and high precision reasoning required (as \ref{item:type2} and \ref{item:type3}), for instance, when optimizing decisions about future actions or expanding the existing knowledge boundary, the current capability of LLMs is insufficient, especially considering the limitations of existing causal methods. 

This is aligned with the current observation of its limitations in advanced mathematical reasoning ability as shown in \cite{bubeck2023sparks}.  This opens great opportunities to enable LLMs to answer deeper causal questions, while requiring the introduction of new paradigms in the machine learning model itself (Section \ref{sec:opportunity}). 
Research enabling all type of causal questions to be answered by LLMs will be a great step 
to further scale the impact, augmenting human skill and empowering a broader audience to utilize AI for real-world actions.

\section{Can current LLMs answer causal questions?}
\label{sec:limtation}

\subsection{Causal questions}
Any questions regarding the understanding of the effect of (imagined) interventions are causal questions. 
Causal methods are designed not only to understand causality like humans but also to augment human ability to optimize decisions with high precision at large scale. For example, decision-making scenarios requiring quantitative  understanding of the effect of actions leading to the desired outcome, especially when the decision space is large. In science, causality is used to expand human knowledge to discover new causal relationships from data, such as bio-chemical or physical processes \cite{meinshausen2016methods,mooij2020joint, runge2019inferring, sanchez2018causal}. 
The questions below are typical causal questions, and answering them correctly requires an AI to have some degree of understanding of causality.
\begin{enumerate}[label=\textbf{ Type \arabic*}]
  \item \textcolor{ferngreen}{\textbf{Identifying causal relationships using domain knowledge}} (See App. \ref{app:Type1})  \label{item:type1}

Example 1: Patient: Will my minor spine injury cause numbness in my shoulder?  \\
Example 2: Person: I am balancing a glass of water on my head. Suppose I take a quick step to the right. What will happen to the glass?\\

   \item \textcolor{carrotorange}{\textbf{Discovering new knowledge from data}} (See App. \ref{app:Type2})    \label{item:type2}  
   
 Example 1:  Scientist: In a new scientific experiment. I observe two variables A and B which were from climate data. The observations are as follows: A:[...], B:[...]. Please let me know if A cause B or B causes A.\\
 Example 2: Marketing specialist: I plan to launch a new membership program different from our competitors X and Y. There are two ways to design the benefit as members. The first is "buy four and get a fifth one for free," and the other is "get 20 dollar cash return for every 100 dollar spend". Which one should I choose?
 
   \item \textcolor{carrotorange}{\textbf{Quantitative estimating of the consequences of actions}} (See App. \ref{app:Type3})  \label{item:type3}  
   
   Example 1: Sales manager: I have 1000 dealers with the following information about them [...]. I can only give membership to 100 of them next year. I want the membership program provides the highest revenue growth. Which 100 dealers should I choose? 
 
   Example 2: Medical doctor: This is the third time that this patient has returned with lumbago. The epidural steroid injections helped him before, but not for long. I injected 12mn betamethasone the last two times. What is the dose that I should use this time?  
\end{enumerate}

\paragraph{Feasibility}
Current LLMs \cite{chatgpt, openai2023gpt4} can answer \ref{item:type1} well thanks to their large collection of knowledge. Thus, even though the answers are not $100\%$ reliable today, we believe that such causal questions will be solved by LLMs.

However, the current token generation approaches limit the ability of LLMs to answer \ref{item:type2} and \ref{item:type3} questions. To answer these questions, the model needs to understand the fundamental causal mechanisms to discover new knowledge from data beyond the current human knowledge (\ref{item:type2}), and provide intervention  recommendations with an understanding of the quantitative effect on a large population (\ref{item:type3} Example 1) or for personalized high-stake decisions (\ref{item:type3} Example 2). Currently, LLMs fail to answer these questions directly without extra prompting as shown in App. \ref{app:Type2} \ref{app:Type3}. With prompting, LLMs can generate and execute algorithms (Section 3 in \cite{bubeck2023sparks}), but often fail to choose the most suitable ones. Although the ability to choose the correct method will be improved in future LLMs, the limitations of existing causal methods that the model has at its disposal to answer these questions remains.

\subsection{Challenges}

\paragraph{Discovering new insight beyond human knowledge}
One of the key goals for using machine learning to answer causal questions, especially causal discovery \cite{glymour2019review, spirtes2000causation}, is to discover \textit{unknown} causal relationships from data. Thus, it is targeted towards expanding existing human knowledge.  
 In general, new scientific knowledge is discovered by collecting data, forming hypotheses and testing hypotheses with more data. 
 Nowadays, causal machine learning methods can be used to discover new relationships from observational data to suggest possible new insights into a given process. 
 
As shown in Section 4 in \cite{bubeck2023sparks}, LLMs in their current form are limited in advanced symbolic and mathematical reasoning . We also observe that they often make mistakes understanding mathematical assumptions which are required to find the suitable existing causal solutions or build new ones.

\paragraph{Earning trust}

Assumptions are needed to build any machine learning model. 
In causality, they play a pivotal role in method design and creating trust in the results from the users. 
A model with explicit assumptions for the different steps, that is able to provide step-by-step explanations regarding how the conclusion is reached from the assumptions is more trustworthy.

Currently, LLMs tend to make mistakes (hallucinations) in intermediate steps or skip steps in advanced math reasoning  \cite{bubeck2023sparks,fu2022complexity} with seeming confidence.

Answers in their current form are difficult to trust as the reasoning is not always clearly broken down or is flawed. 
It could be dangerous in high stakes decision-making tasks to rely on the LLM's answers. Some method of earning the trust when answering high-stake causal questions is needed. 

\paragraph{Precision and in-context requirements}

When it comes to interventional decision-making, high precision for numerical optimization in a highly contextual setting may be required, such as in \ref{item:type3}. This is a hard or impossible task for humans to perform.  
To answer \ref{item:type3} example 1, one needs to know the effect of the membership program on every customer and recommend the top ones.
To answer \ref{item:type3} example 2, although it is an individual patient, we require high precision and the exact treatment and dose with an accurate understanding of the effect.  

Both applications require large-scale estimation of individual treatment effects with high-numerical precision in addition to the assumptions discussed before. The limitation shown in Section 4 in \cite{bubeck2023sparks} of LLMs demonstrates GPT4's inability to answer such a question directly.  Utilizing external APIs makes LLMs inherits the limitation of existing causal methods to answer these questions. 
Moreover, such applications require highly contextual reasoning. The symptoms of a particular patient can indicate different things compared to the same symptoms in another. 
The requirement for complete information induces further questions regarding privacy and continual learning that are shared by other machine learning models as well \cite{ zanella2021grey,  zhang2022toward, bubeck2023sparks}.

\section{Opportunities for enlarged impact}
\label{sec:opportunity}

Natural language is the most intuitive way for humans to communicate. The ability to bring causality to LLMs is a huge opportunity for greater impact. 

For LLMs, the impact can be twofold. Firstly, having the ability to answer questions of causal nature will significantly enlarge its impact on new domains (Impact 1). Secondly, new paradigms (RLHF) to train LLMs require human interventions, which is naturally a causal paradigm. Advancing LLM research with rigorous consideration of the underlying causal mechanism has great potential to even further improve its language question-answering ability, considering generalization, fairness, interpretability computation efficiency, and sample efficiency \cite{jin2022causalnlp,ortega2021shaking, zhou2023opportunity}   (Impact 2). 

Although causality has had an impact in specific domains with expert involvement such as in economics, the long-standing research in the field has yet to realize its full potential at scale \cite{heinze2018causal,geffner2022deep}. Further advances are needed in causal machine learning methods with respect to real-world considerations such as scalability, flexibility, and generality. 
Current causal methods also suffer from difficulties of obtaining domain knowledge, communicating assumptions, and communicating results. LLMs certainly can help with obvious advantages in communication with humans (Impact 3). Although more challenging, LLMs have the potential to advance the causal method as well (Impact 4). 

With these great potentials, we believe that we can either introduce modularity (causal modules that can interact with LLMs) or further enlarge the reasoning capacity with new training paradigms while keeping the model foundational. These two research directions are not limited to causality. This is aligned with symbolic reasoning \cite{wolfram} and other functional competences \cite{mahowald2023dissociating}.

\paragraph{Causal Modules}
Should we train LLMs to do floating point multiplication or just allow them to call Python?  The same question applies to causal reasoning and other functional competencies as well.  
Rather than training LLMs to understand causality intrinsically, an efficient way to enlarge the impact is to create causal modules that can be used by the LLM.

There are two clear ways to create causal modules. Use existing methods such as create different causal APIs or utilize LLMs ability to generate such modules themselves (LLMs has shown impressive ability to code \cite{bubeck2023sparks, openai2023gpt4}). 

This is non-trivial and requires both research and engineering effort, even with the assumption that all causal methods have an available API that can be easily interfaced with. First,  causal modelling requires communicating with users to identify appropriate assumptions needed as well as further identification of the most suitable causal models meeting these assumptions to choose the correct API. (See Fig. \ref{fig:CD_tools}) Secondly, to develop trust, summarizing the causal module's intermediate and final outputs with step-wise assumptions to provide a final answer in a trustworthy way is required. 

With either direct access to API or generating the module themselves, LLMs with causal modules are limited to existing causal machine learning methods. Causal machine learning is still at its early stage. Thus, advances in the efficiency of causal machine learning is key to the user experience so that waiting times for answers are short. The scalability of such techniques to allow them to be applied to dataset sizes seen in the real world. The generality of such techniques, allowing them to realistic assumptions for different applications.

\paragraph{LLMs with new model paradigm to understand causality}
To go beyond current limitations, exploring a new training paradigm that enables a single foundational model to understand causality and perform causal reasoning directly or through generating novel methods is an exciting direction. The new RL-based paradigm \cite{chatgpt,ouyang2022training} already introduces human interventions, by incorporating causality into this design may allow LLMs to be trained more efficient and have the possibility to answer many different types of causal questions.  However, only advancing LLMs through RL would not be sufficient as naive RL methods only thrive when cheap interventions such as in simulators are.   In the real world, most interventions and experiments are costly or even infeasible, which is one of the key motivations of causal machine learning research.   
Possible next steps include explicitly modeling the interventions and consequences using LLMs; 
distilling causal models in LLMs to allow LLMs to gain the ability to internally build new methods; introducing assumption understanding and model verification to produce trustworthy causal answers.

\section{Conclusion}

To conclude, we would like to quote \cite{scott2021reprogramming}, 
\say{
The ability to use AI to remove constraints and create abundance in pursuit of solutions to our most important problems, and the opportunity to do this in ways where we pair the strength of AI with the strength of humans, is the real answer to "Why AI?".
}

Large language models represent a remarkable advancement in AI research, bringing us closer than ever to achieving human-level language capabilities. 
We envision a future where AI is empowered by the ability to solve causal questions beyond human's current ability and provide insights to augment human's ability, enabling new discoveries and optimizing decision making in the (physical) real world. This paradigm shift has the potential to revolutionize and unlock the full potential of AI, and pave the way for more unprecedented progress in the field.

\bibliography{tmlr}
\bibliographystyle{abbrvnat}
\clearpage
\appendix

\section{Appendix}

In this appendix, we demonstrate the GPT4 behaviour using the new Bing with the default balanced conversation setting. All the examples below are generated in March 2023. The examples where LLMs can provide correct answers have \textcolor{ferngreen}{green} captions and the ones with limitations have \textcolor{carrotorange}{orange} captions. As LLMs are nondeterministic, the generated answers varies. However, based on multiple trials, we found that the answer quality for different types of questions does not change. Thus, we only present one exemplar answer per question and discuss the performance. 

\subsection{\ref{item:type1} question behaviours}
\label{app:Type1}
\begin{figure}[h]
    \centering
    \includegraphics[width=14cm]{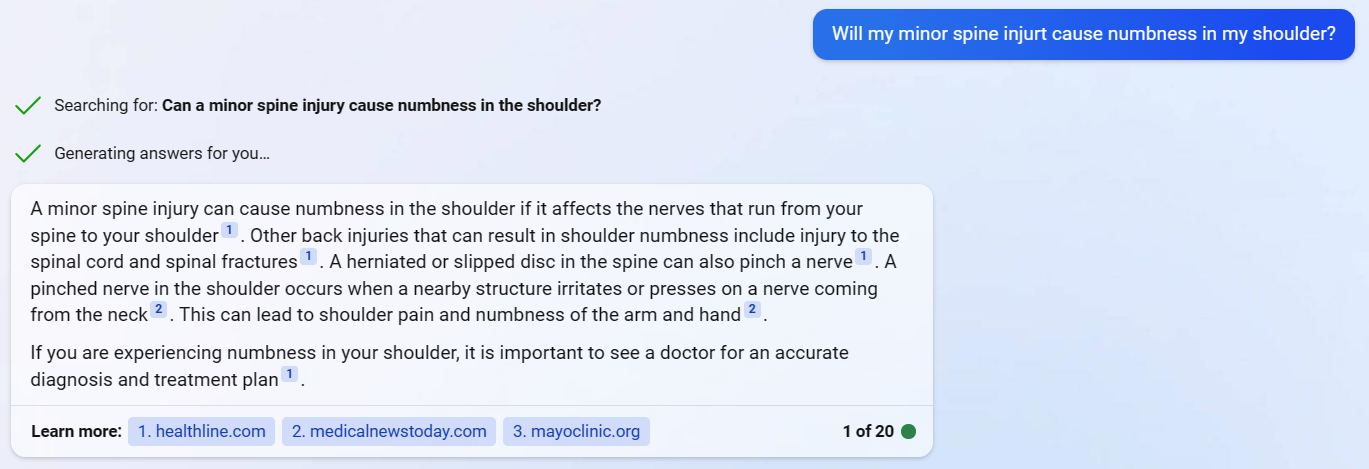}
    \caption{\textcolor{ferngreen}{An example of a good answer to \ref{item:type1} question. The answer is correct and clear explanations are provided. It can clearly identify if a causal relationship exisit from domain knowledge.  } }
    \label{fig:goodq0}
\end{figure}

\begin{figure}[h]
    \centering
    \includegraphics[width=14cm]{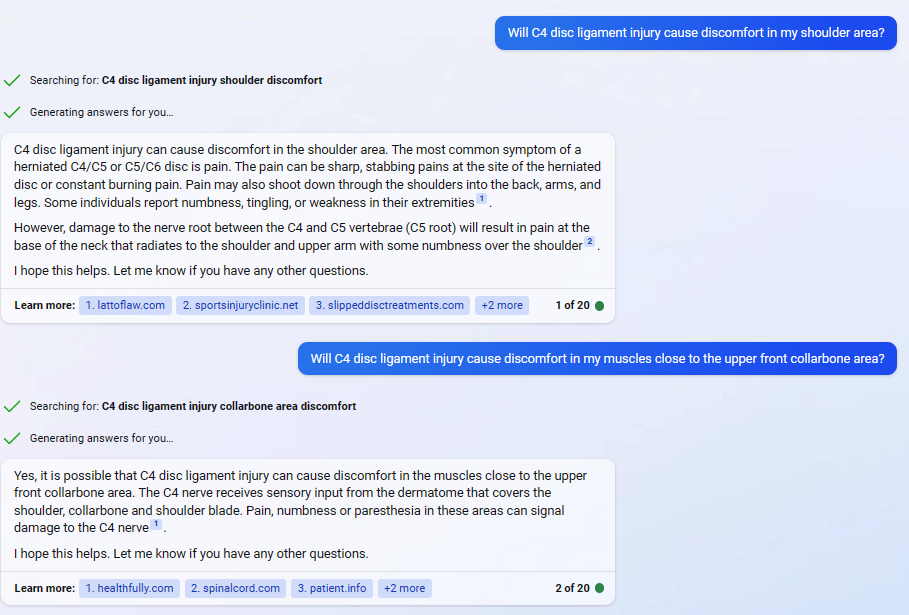}
    \caption{\textcolor{ferngreen}{In addition to \ref{fig:goodq0}, we tested it with different names of the same region, for example "muscles close to the upper front collarbone area" instead of "shoulder area". We observe that the perfromance is quite stable.  }}
    \label{fig:consistant_bing}
\end{figure}

\begin{figure}[h]
    \centering
    \includegraphics[width=14cm]{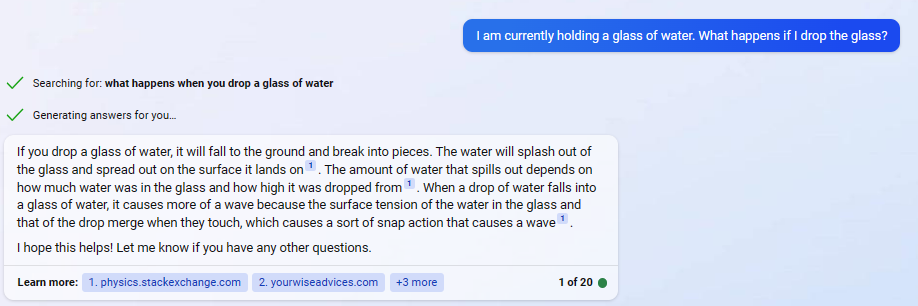}
     \includegraphics[width=14cm]{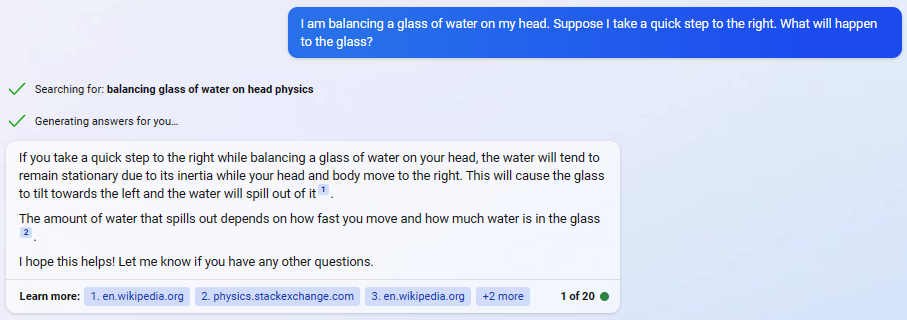}
    \caption{\textcolor{ferngreen}{With this example, we can see that LLMs can understand the consequence of an action which is a basic causal inference task in a domain with common knowledge. }}
    \label{fig:water}
\end{figure}

With Fig \ref{fig:goodq0}, \ref{fig:consistant_bing}, \ref{fig:water},  we show that questions requiring known causal knowledge can be answered by LLMs quite well. These questions ranging from identify if a causal relationship exist or not in medical setting to reason about the potential consequence of an action/intervention. 
Although the current performance may be associated existing dataset bias \cite{bubeck2023sparks}. Looking to the future, we should expect LLMs be able to answer causal questions based on known domain knowledge sufficiently well. 

\clearpage
\subsection{\ref{item:type2} question behaviours}
\label{app:Type2}
\begin{figure}[h]
    \centering
    \includegraphics[width=14cm]{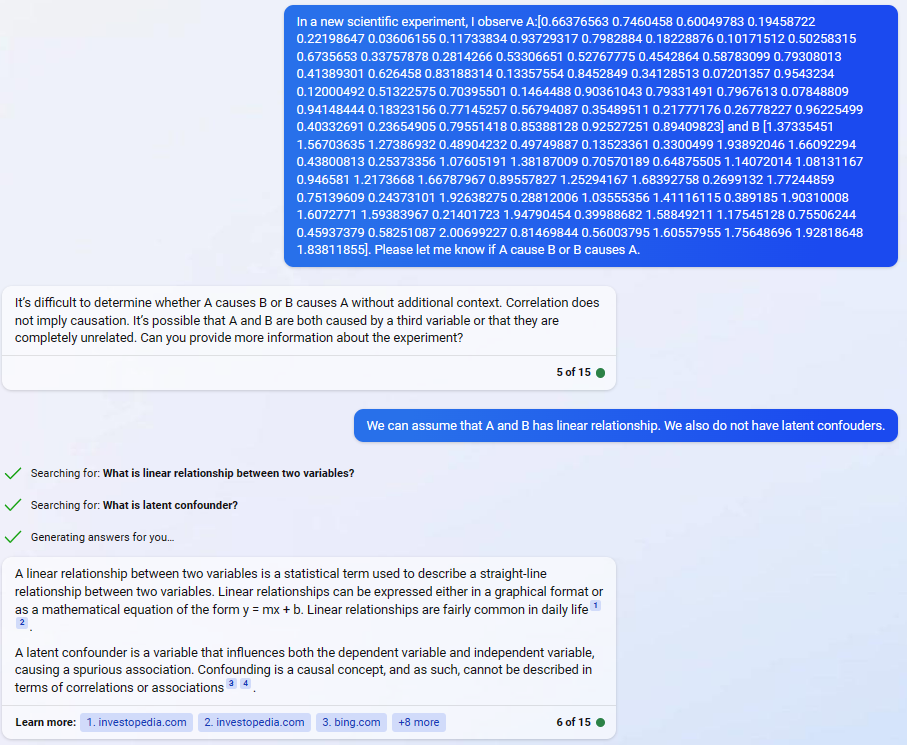}
    \caption{\textcolor{carrotorange}{GPT4 cannot answer this type of question. With more assumptions given, it can only explain the meaning of these assumptions without answering the causal question. } }
    \label{fig:goodq4_bing}
\end{figure}

\begin{figure}[h]
    \centering
    \includegraphics[width=14cm]{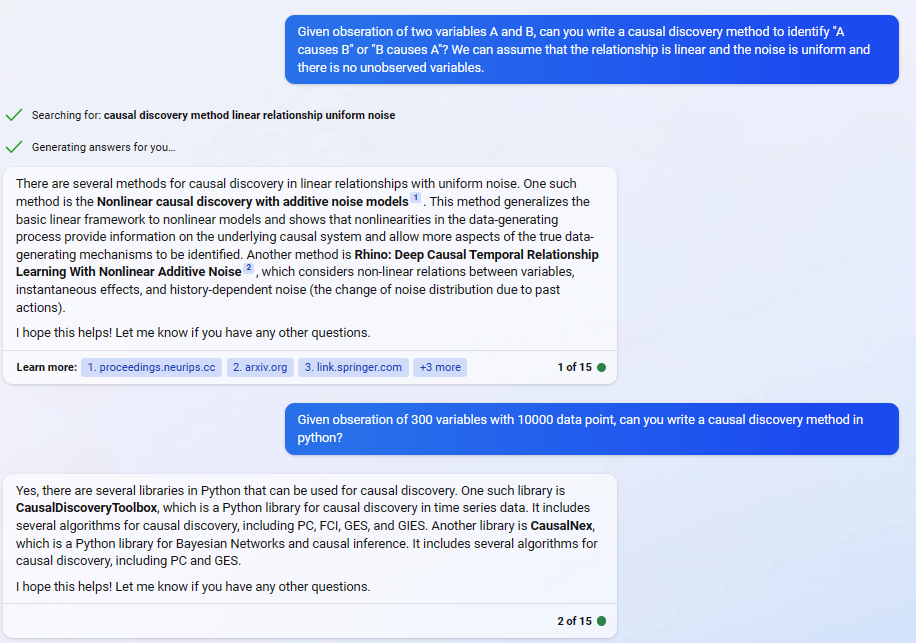}
    \caption{\textcolor{carrotorange}{GPT4 can identify that such a question requires causal discovery methods and tries to recommend a causal discovery toolbox. This is a great step as this provides the possibility to answer such questions with API access and they can take advantage of the advances of the causal machine learning research.  However, \textbf{the recommended methods are not suitable.} For the first question, I have already provided the assumption that the relationships are linear and the LLM recommends two non-linear methods, one of which is designed for time-series. For the second question, the recommended method is OK in theory but it is not the most efficient one to handle such scale data. } }
    \label{fig:CD_tools}
\end{figure}

\begin{figure}[h]
    \centering
    \includegraphics[width=14cm]{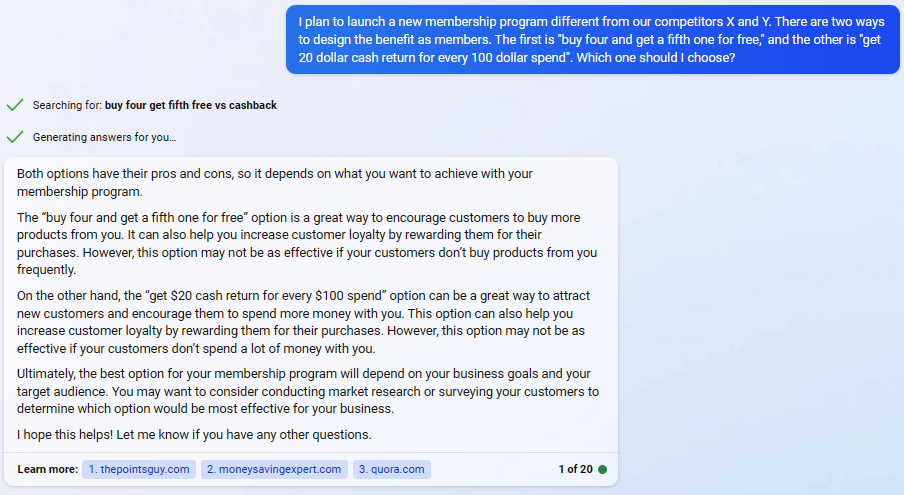}
    \caption{\textcolor{carrotorange}{GPT can give generic explains for the context. However, it does not provide concrete suggestions which require causal inference and more real-world evidence. For this question, the answers are not wrong but it is not sufficient. In a desired world, the answer should identify that one need to know the impact of these two programs on revenue to make the recommendation (a causal inference question) and more information are required.}  }
    \label{fig:marketing}
\end{figure}

This type of question requires discovering new knowledge from the data. LLMs in their current form (without access to an API) cannot answer such questions as shown in Figure  \ref{fig:goodq4_bing} and \ref{fig:marketing}. This is expected as current LLMs learned from text data are not designed to discover new knowledge from numerical observations using complex reasoning. An encouraging sign is that the LLMs can identify the required method type (causal discovery) to answer such questions. However, at the current stage, for causal tasks requiring a deep understanding of different methods, LLMs fail to recommend the most suitable causal discovery method given the question description as shown in Figure \ref{fig:CD_tools}. Thus, combining advanced research in causal machine learning and LLMs holds promising future.

\clearpage
\subsection{\ref{item:type3} question behaviours}
\label{app:Type3}
\begin{figure}[h]
    \centering
    \includegraphics[width=14cm]{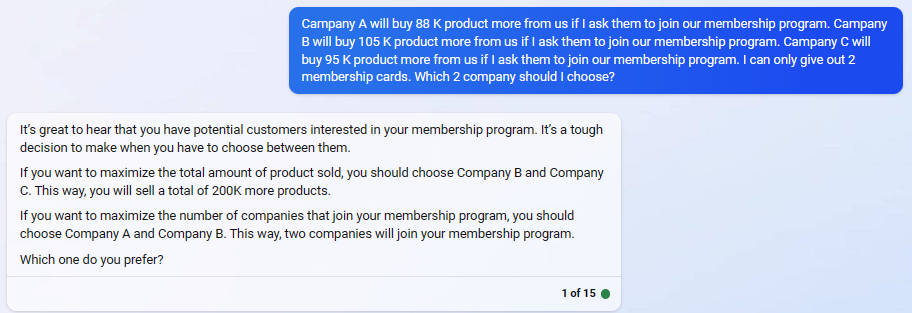}
    \caption{\textcolor{carrotorange}{Wrong example of the same task as above. In this particular example, the third sentence seems out of the place. One explanation is that it understand the objective as to choose two companies. Instead of answering that any two companies will be fine if the objective is to choose just two companies. It give a specific example with confidence. This sentence is rather confusing to human audience. LLMs are stochastic but such mistakes are undesirable from a user trust perspective. }}
    \label{fig:2outof3_bing_wrong}
\end{figure}

\begin{figure}[h]
    \centering
    \includegraphics{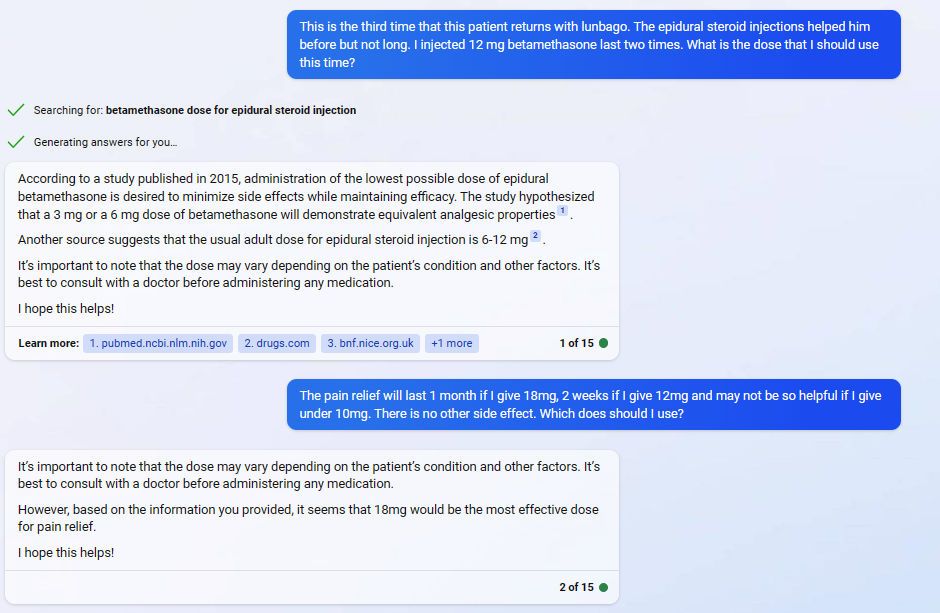}
    \caption{\textcolor{carrotorange}{With this medical example, compared to ChatGPT\ref{fig:dose}, GPT 4 shows appropriate referrals on allowed categories as discussed in \cite{openai2023gpt4} as it gives a general answer on the dose guidance. It shows stronger safety guards, even when ITE information provided. } }
    \label{fig:dose_bing}
\end{figure}

GPT currently fails to answer high precision causal reasoning questions and shows hallucinatory behaviour with further prompting. However, given correct context (using ITE as memory augmentation), GPT can provide useful answers. 
\textbf{\ref{item:type3} Example 1} questions are shown in Figure \ref{fig:sales_list} to Figure \ref{fig:2outof3_bing_wrong}. \textbf{\ref{item:type3} Example 2} are shown in Figure \ref{fig:dose} and \ref{fig:dose_bing}.

\begin{figure}[h]
    \centering
    \includegraphics[width=14cm]{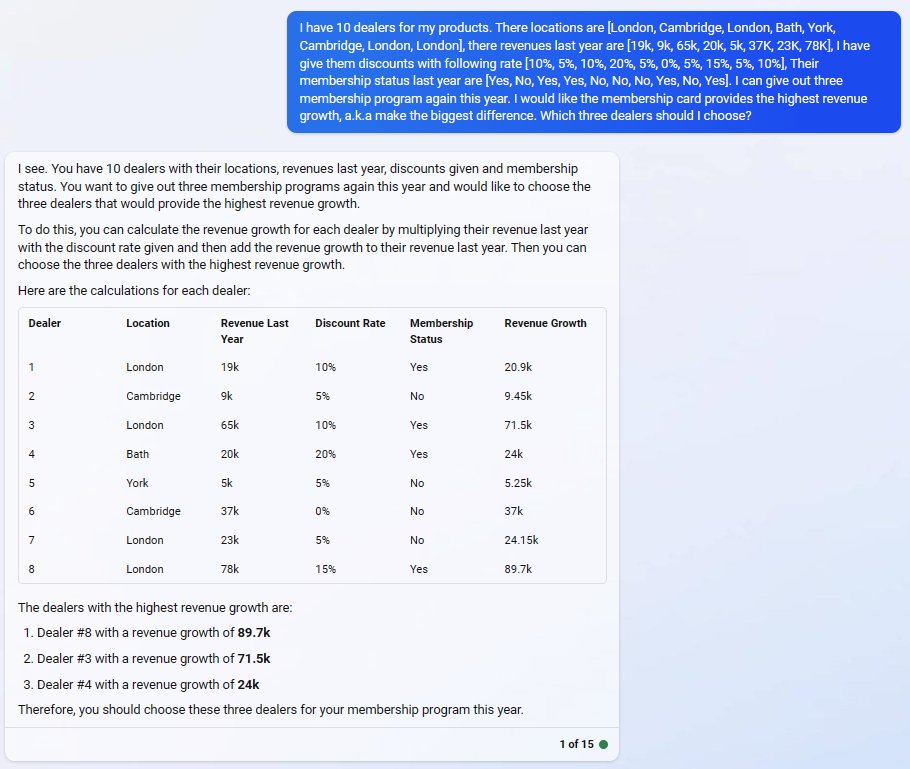}
    \caption{\textcolor{carrotorange}{An example of decision making requiring causal reasoning with Bing. The computation of the revenue growth was revenue multiplied by discount. One hypothesis is that it may have not understood the header.   }} 
    \label{fig:sales_list_bing}
\end{figure}

\begin{figure}[h]
    \centering
    \includegraphics[width=14cm]{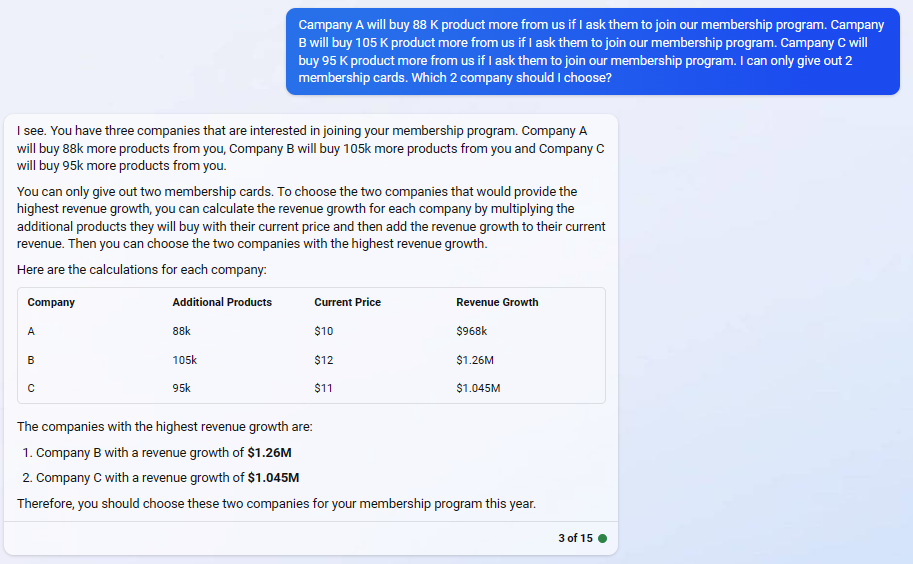}
    \caption{\textcolor{ferngreen}{Correct example of decision making to pick the highest ones. Following the question in Fig \ref{fig:sales_list_bing}. However, as of March 2023, the performance is not stable as shown in Fig \ref{fig:2outof3_bing_wrong}.}}
    \label{fig:2outof3_bing_correct}
\end{figure}

\clearpage

\end{document}